\documentclass[12pt]{article}
\usepackage[left = 1 in, right = 1 in, top = 1 in, bottom = 1.5 in]{geometry}              
\usepackage[utf8]{inputenc}
\usepackage{amsmath, amsthm, amssymb,xcolor,multicol,url}
\usepackage[sectionbib, round]{natbib}
\usepackage{graphicx,bbm}
\usepackage{booktabs,epstopdf,color}
\usepackage[space]{grffile}
\usepackage{lineno,comment}
\usepackage[plain,noend]{algorithm2e}
\usepackage{multirow}

\usepackage{hyperref}

\usepackage{amssymb}
\usepackage{subfig}

\usepackage{authblk}

\makeatletter
\renewcommand{\algocf@captiontext}[2]{#1\algocf@typo. \AlCapFnt{}#2} 
\def\@algocf@capt@plain{top}
\renewcommand{\algocf@makecaption}[2]{%
  \addtolength{\hsize}{\algomargin}%
  \sbox\@tempboxa{\algocf@captiontext{#1}{#2}}%
  \ifdim\wd\@tempboxa >\hsize
    \hskip .5\algomargin%
    \parbox[t]{\hsize}{\algocf@captiontext{#1}{#2}}
  \else%
    \global\@minipagefalse%
    \hbox to\hsize{\box\@tempboxa}
  \fi%
  \addtolength{\hsize}{-\algomargin}%
}
\makeatother

\def\T{{ \mathrm{\scriptscriptstyle T} }}

\DeclareMathOperator*{\argmin}{arg\,min}
\DeclareMathOperator*{\argmax}{arg\,max}

\def\T{{\mathrm{\scriptscriptstyle T} }}

\newcommand{\be}{\begin{equs}}
\newcommand{\ee}{\end{equs}}

\numberwithin{equation}{section}

\newtheorem{definition}{Definition}

\begin{document}
\title{Constrained Reweighting of Distributions: an Optimal Transport Approach}

\author{Abhisek Chakraborty, Anirban Bhattacharya, Debdeep Pati\\
Department of Statistics, Texas A\&M University,\\ College Station, TX 77843, U.S.A.
}

\maketitle

\begin{abstract}
We commonly encounter the problem of identifying an optimally weight adjusted version of the empirical distribution of observed data, adhering to predefined constraints on the weights. Such constraints often manifest as restrictions on the moments, tail behaviour, shapes, number of modes, etc., of the resulting  weight adjusted empirical distribution.  In this article, we substantially enhance the flexibility of such methodology by introducing a nonparametrically imbued distributional constraints on the weights, and developing a general framework leveraging the maximum entropy principle and tools from optimal transport. The key idea is to ensure that the maximum entropy weight adjusted empirical distribution of the observed data is close to a pre-specified  probability distribution in terms of the optimal transport metric while allowing for subtle departures. The versatility of the framework is demonstrated in the context of three disparate applications where data re-weighting is warranted to satisfy side constraints on the optimization problem at the heart of the statistical task: namely, portfolio allocation, semi-parametric inference for complex surveys, and ensuring algorithmic fairness in machine learning algorithms. 

\end{abstract}

\section{Introduction}
Maximum entropy principle \citep{shannon1948mathematical,jaynes1957information} states that in situations characterized by uncertainty and limited prior-knowledge-guided constraints, the optimal choice among all feasible probability distributions is the probability distribution that is the least informative or most uniformly spread. This idea is at the heart of numerous statistical tasks that has permeated into every corner of modern machine learning research. Prominent instances of such constrained entropy maximization include applications in  image reconstruction \citep{skilling1984maximum}, ill-posed inverse problems \citep{gamboa1997bayesian}, portfolio optimization \citep{doi:10.1080/07474930801960394}, generalised methods of moment models \citep{Chib2018}, natural language processing \citep{gudivada2018computational}, network analysis  \citep{magrans2018connectivity}, reinforcement learning \citep{DBLP:journals/corr/abs-2103-06257}, to name a few. We refer the readers to \citet{kardar2007statistical, cover2012elements} for book-length reviews. 

For maximum entropy inference, the specified constraints imposed on the probability distributions frequently manifest as constraints pertaining to moments \citep{Chib2018}, tail characteristics \citep{Bernoulli14-4-2008}, distributional shapes \citep{Chernozhukov2023}, modal counts, and similar properties. In many cases, however, constructing constraints with the desired level of flexibility is challenging, if not unfeasible -- refer to Sections \ref{ssec:cx_surveys} and \ref{ssec:po} for specific examples in the context of inference in complex surveys and moment condition based of portfolio optimization respectively. On a related note, a recent article \citep{chakraborty2023robust} introduced a flexible framework for introducing more elaborate constraints on probability distributions in the context of conducting robust Bayesian inference.

In this article, we offer a novel solution to this problem via introducing a  probability distribution-guided constrained entropy maximization framework, that not only offers versatility but also enhances the  interpretability of the inferential output. The main concept revolves around ensuring that a weight adjusted empirical distribution of the observed data closely aligns with a predetermined family of probability distributions, measured through a statistical distance \citep{rachev2007advanced}. Importantly, the family of probability distributions is potentially continuous, but any weighted-adjusted empirical distribution of the observed data is discrete. This eliminates the possibility of adopting many common statistical discrepancies, e.g Kullback-Leibler, total variation, Hellinger's distance, to place the probability distribution-guided constraints. In practice,  we need to exercise ardent care for our choice tailored to the application of interest. For homogeneity of exposition across all scenarios in this article, we considered the Wasserstein metric  \citep{Villani2003TopicsIO, noauthororeditor}. 

The idea of data re-weighting is of course not new. \cite{wang2017robust} suggested elevating the likelihood of individual observations using data-driven weights, to conduct robust inference under mild  model misspecification. \cite{wen2014robust} proposed a data re-weighting scheme to align the data with a different target distribution, enabling inference under covariate shift. Other compelling ideas involving re-weighting can be traced in  fair learning \citep{yan2022forml}, natural language processing \citep{GarridoRamas2022}, variational tempering \citep{mandt2016variational}, etc. Complementing the existing literature, we propose a versatile data re-weighting framework, borrowing from  the maximum entropy principle and optimal transport, that renders itself useful in a multitude of statistical tasks.

The rest of the paper is organised as follows. The general framework of the proposed probability distribution guided constrained  entropy  maximization is motivated and introduced in Section \ref{sec:method}. Section \ref{ssec:cx_surveys},  \ref{ssec:algorithmic_fairness}, and \ref{ssec:po} presents   applications of our methodology in the context of  semi-parametric inference in complex surveys,  in ensuring demographic parity in machine learning algorithms, and entropy based portfolio optimization respectively. Finally, we conclude with a discussion.

\section{General Framework}\label{sec:method}

Let $[a]$ denote the set of integers $\{1, \ldots, a\}$. Let $\Omega$ denote the set of all possible discrete distributions $\omega$ with atoms $\mathbf{s} =(s_1, \ldots, s_m)^{\T}$. The entropy of the discrete probability distribution $\sum_{i=1}^m w_i\delta_{s_i}(\cdot)$ is defined by 
\begin{align*}
   \mbox{H}_m(\mathbf{w}) = -\sum_{i=1}^m w_i\log w_i, 
\end{align*}
where $\delta$ is the Dirac's delta function. The entropy $\mbox{H}_m(\mathbf{w})$ is  a measure of randomness which is maximized at the discrete uniform distribution with $w_i = 1/m$ for all $i$.
In many statistical tasks, the core challenge constitutes of optimizing  a functional
$\mathcal{F}:\Omega\to\Omega^{\prime}$  with respect to $\omega$ subject to a constraint $\omega\in\Omega_0 (\subset\Omega)$. A simple example is  when $\mathbf{s} = (s_1, \ldots, s_m)^{\T}$ is the observed sample itself. Then, the set $\Omega$ is simply characterised by the class of weighted empirical distributions of the observed data,
\begin{align*}
    \Omega =  \big\{\omega=\sum_{j=1}^m w_j\ \delta_{s_j}(\cdot): \sum_{j=1}^m w_j = 1,\ w_j\geq 0,\ j\in[m]\big\}.
\end{align*}
In the sequel, we shall see more general examples where the constraint set $\Omega_0$ can be identified with a subset of an $(m-1)$-dimensional probability simplex $\mathcal{S}_{m-1} = \{\mathbf{w}: \sum_{i=1}^m w_i = 1,\ w_i>0,\ i\in[m]\}$, for some $m\in\{1,2,\ldots\}$.

Given $\mathbf{s} = (s_1, \ldots, s_n)^{\T}$, parametric inference constitutes approximating the empirical distribution $(1/n)\sum_{i=1}^n \delta_{s_i}(\cdot)$  via a parametric family of distributions $\{f_\theta: \theta \in \Theta\}$, and learn the parameter $\theta$ from data. Such procedures often fall prey to model misspecification \citep{5805f73c-4dfa-385e-bd6d-68424fb9f5be}, leading to untrustworthy inference. To avoid complete model specification,  a popular class of semi-parametric approaches \citep{hall2005generalized} operate under a milder  assumption  that the weight adjusted empirical distribution $\sum_{i=1}^n w_i \delta_{s_i}(\cdot)$ satisfies  moment restrictions of the form $\sum_{i=1}^n w_i\ g(s_i, \theta) = 0$, where  $g$ is vector of known functions on $\mathbf{R}^d\times\Theta$. In numerous instances, achieving such moment based constraints with the intended degree of flexibility proves to be arduous, if not practically impossible--we elaborate on this more in the sequel.
To that end, in this article, we offer a middle ground between the fully parametric and semi-parametric moment condition models, that allows for flexible modeling assumptions while enjoying coherent interpretability similar to parametric inference. We propose to operate under a restriction of the form $\mbox{D}(\sum_{i=1}^n w_i \delta_{s_i}(\cdot), f_{\theta})\leq \varepsilon$, where $\mbox{D}$ is a statistical distance and $\varepsilon$ is an user-defined hyper-parameter. Our goal is to infer about $\theta$ while allowing for  mild deviations from the parametric model $f_\theta$, and $\varepsilon$ measures the maximum allowable discrepancy. 


Inference under moment condition models often constitutes computing the maximum entropy weighted-adjusted empirical distribution of $(s_1, \ldots, s_n)^{\T}$ that satisfies some pre-specified moment conditions \citep{Chib2018, chib2021bayesian}. That is, for every $\theta\in\Theta$, we calculate $\sum_{i=1}^n w^{\star}_i(\theta)\ \delta_{s_i}(\cdot)$  where  $\mathbf{w}^{\star}(\theta) = \argmax_{\mathbf{w}\in \mathcal{S}_{n-1}} \mbox{H}_n(\mathbf{w})$, subject to $\sum_{i=1}^n w_i g(s_i, \theta) = 0$.
Under the proposed framework, we too appeal to the maximum entropy principle and compute the maximum entropy weight adjusted empirical distribution of $\mathbf{s}$ that satisfies the parametric distribution guided constraint. That is, for every $\theta\in\Theta$, we calculate $\sum_{i=1}^n w^{\star}_i(\theta)\ \delta_{s_i}(\cdot)$  where
\begin{align}
  &\mathbf{w}^{\star}(\theta) = \argmax_{\mathbf{w}\in \mathcal{S}_{n-1}} \mbox{H}_n(\mathbf{w})\ \text{subject to}\ \mbox{D}\bigg(f_{\theta},\ \sum_{i=1}^n w_i\ \delta_{s_i}(\cdot)\bigg) \leq \varepsilon,
\end{align}
where $\mbox{D}$ is a statistical distance, and $\varepsilon$ is an user defined parameter. In ensuing applications in this article, we often solve the dual optimization problem for operational ease. In that, for each $\theta\in\Theta$ and $\lambda\geq 0$, we calculate $\sum_{i=1}^n w^{\star}_i(\theta)\ \delta_{s_i}(\cdot)$  such that 
\begin{align}
  \mathbf{w}^{\star}(\theta) = \argmax_{\mathbf{w}\in \mathcal{S}_{n-1}}\bigg[\mbox{H}_n(\mathbf{w}) -\lambda\ \mbox{D}\bigg(f_{\theta},\ \sum_{i=1}^n w_i\ \delta_{s_i}(\cdot)\bigg)\bigg].
\end{align}
The parameter $\lambda$ controls the extent of departure from the guiding parametric distribution. 

One pivotal aspect yet to be addressed within the proposed framework is that, an weight adjusted empirical distribution is discrete, but in the context of a specific problem, the guiding distribution $f_{\theta}$ is potentially continuous. For instance, in Section \ref{ssec:po}, in the context of entropy based portfolio allocation, $f_{\theta}$ takes the form of a skew normal distribution \citep{SkewNormal}. This precludes the utilization of several standard statistical distances, such as total variation, Hellinger's distance, $\chi^2$ distance, etc., for implementing the distance-based constraint.  In this article, due to its versatility, we employ the Wasserstein metric \citep{Villani2003TopicsIO} with $L_2$ cost as the distance measure $\mbox{D}$.  To that end, we briefly recall some relevant facts about the $2$-Wasserstein metric.  The Wasserstein space $\mathbf{P}_2(\mathbf{R}^d)$ is defined as the set of probability measures $\mu$ with finite moment of order $2$, i.e $\{\mu \,: \, \int_{\mathbf{R}^d} \left\lVert x\right\rVert^2 d\mu(x) < \infty\}$, where $\left\lVert \ \cdot \ \right\rVert$ is the euclidean norm on $\mathbf{R}^d$.

\begin{definition}\label{def_wp}
For $p_0, p_1 \in \mathbf{P}_2(\mathbf{R}^d)$, let $\pi(p_0, p_1) \subset \mathbf{P}_2(\mathbf{R}^d\times\mathbf{R}^d) $ denote the subset of joint probability measures (or {\em couplings}) $\nu$ on $\mathbf{R}^d\times\mathbf{R}^d$ with marginal distributions $p_0$ and $p_1$, respectively. Then, the $2$-Wasserstein distance $W_2$ between $p_0$ and $p_1$ is defined as 
$W_{2}^{2}(p_0, p_1)
= \inf_{\nu\in \pi(p_0,p_1)} \int_{\mathbf{R}^d\times\mathbf{R}^d}\
\left\lVert y_0 - y_1 \right\rVert^2\ d\nu(y_0,y_1)$.
\end{definition}
Importantly, if both $p_0, p_1\in \mathbf{P}_2(\mathbf{R})$ with quantile functions $F^{-1}_0, F^{-1}_1$, we have an easily tractable expression, \citep{2019}
  $W_{2}^{2}(p_0, p_1) = \int_{[0, 1]} \big[F^{-1}_0(q) - F^{-1}_0(q)\big]^2 dq.$
This is heavily utilized  in subsequent sections.  With that, we have all the essential ingredients to delve down on the specific applications of interest.

An instance of application of the proposed framework emerges within the realm of semi-parametric inference in complex survey data \citep{anzstat_cxsurvey, nhanes}.
In survey sampling, we wish to infer about a collection of features of a finite population $\mathcal{P}:=\{X_i,\ i\in[N]\}$. We are provided with a non-representative sample $(x_1, \ldots, x_n)$ obtained from $\mathcal{P}$ via a complex survey, and the corresponding survey weights $\pi = (\pi_1, \ldots, \pi_n),\ 0 < \pi_i < \infty$. In the general framework, this task involves finding the optimal $\omega\in\Omega_0\subset\Omega$ such that 
\begin{align*}
 \Omega =\big\{\omega=\sum_{j=1}^n w_j\delta_{s_i}(\cdot): \sum_{j=1}^n w_j = 1, w_j\geq 0,\ j\in[n]\big\},   
\end{align*}
where $(s_1,\ldots, s_n) = (x^{\star}_1,\ldots, x^{\star}_n)$ is an i.i.d pseudo sample of size $n$ obtained from the complex survey sample $(x_1, \ldots, x_n)$, via weighted finite population Bayesian bootstrap \citep{dong2014a, cohen1997bayesian,  lo1993bayesian} to adjust for the survey weights; and $m = n$. The restriction $\Omega_0$ is dictated by the parametric model that the analyst posits on finite population $\mathcal{P}$ to infer about the features of interest in the finite population. 

The next application in this article deals with the issue  of ensuring demographic parity \citep{90450a4b5b49471b8111fc88355f2e7f, gajane2018formalizing} in machine learning algorithms. Suppose we have data $(x_i, y_i, a_i)\in \mathcal{X}\times\mathcal{Y}\times\{S,T\}$ for  $n$ individuals on  covariate $x\in\mathbf{R}^p$,  continuous response $y\in\mathbf{R}$, and  \emph{protected/sensitive} attribute $A$ with labels $\{S,T\}$. For the sake of simplicity in exposition, we  further assume that $a_i = S,\ i\in[n_S]$, $a_i = T,\ i\in[n]\setminus[n_T]$ and $n = n_S + n_T$.  The goal is to learn a predictive rule $h: \mathcal{X}\times\{S,T\}\to\mathcal{Y}$, that satisfies a specific notion of demographic parity. Refer to section \ref{ssec:algorithmic_fairness} for details. We shall see that, this task involves finding the optimal $\omega\in\Omega_0\subset\Omega$ such that 
\begin{align*}
   \Omega = \big\{\omega=\sum_{j=1}^{n_T} w_{n_S +j}\ \delta_{s_j}(\cdot): \sum_{j=1}^{n_T} w_{n_S+j} = 1, w_{n_S + j}\geq 0,\ j\in[n_T]\big\}, 
\end{align*}
where $s_j = -L(\theta_{(T)}\mid x_j),\ j\in[n]\setminus[n_S]$ is the negative of the loss function utilised to learn the predictive rule $h$ for individuals with $a=T$, $\theta_{(T)}$ are the associated parameters. The optimality of $\omega$ and restriction $\Omega_0$  are determined by the notion of demographic parity utilized. 

An application of a slightly modified version of the general framework is identified in portfolio allocation problems \citep{10.2307/2975974, doi:10.1080/07474930801960394, elton2014modern}, where the goal is  to identify the optimal atoms of the discrete distributions, rather than the weights assigned to the atoms.
This task translates to finding the optimal $\omega\in\Omega_0\subset\Omega$ such that
\begin{align*}
 \Omega   = \big\{\omega=\frac{1}{n}\sum_{i=1}^n \delta_{s_i}(\cdot): \sum_{j=1}^d w_j = 1, w_j\geq 0,\ j\in[d]\big\},   
\end{align*}
where $s_i = \sum_{j=1}^d w_j r_{i, j},\ i\in[n]$; refer to section \ref{ssec:po} for details. The optimality criterion  and the restriction $\Omega_0$ are driven by the fund manager's portfolio allocation objectives and the assumed model for the return distribution, respectively.

\section{Semi-parametric Inference in Complex Surveys}\label{ssec:cx_surveys}

Survey data \citep{anzstat_cxsurvey, nhanes} commonly arises from complex sampling methods such as stratification and multistage sampling wherein individuals in the finite population has unequal probabilities of inclusion into the sample.  Prominent instances of extensive surveys implementing these methodologies include the National Health and Nutrition Examination Surveys (NHANES), the British Household Panel Survey (BHPS), the Household Income and Labour Dynamics in Australia (HILDA) survey, etc.  In complex surveys, the survey sample lacks representativeness, since the individuals with varying demographic characteristics in the finite population of interest,  have varying probabilities of selection into the sample. Consequently, traditional methods of inference and estimation manifests in bias and poor coverage of estimators.

A prevalent approach to address this challenge entails carefully exploiting the sampling weights available with complex survey data sets. These weights could be used to rectify the biases introduced by the unequal probability sampling, and enable us to create pseudo equal probability samples from the population.  If a survey participant falls within a demographic group with a low probability of selection or response, their weight is increased accordingly. Commonly, the available information only includes the survey data set and the associated sampling weights for each unit in the sample. That is, there is limited or no information available about the complex sampling methodology or the precise technique employed for deriving these weights. This situation presents a compelling inferential challenge, which we shall delve into further in the following discussion.

Assume we have a finite population $\mathcal{P}:=\{X_i,\ i\in[ N]\}$, and we wish to infer about a collection of features of $\mathcal{P}$. We are provided with a non-representative sample $x = (x_1, \ldots, x_n)$ obtained via a complex survey design, and the corresponding survey weights $\pi = (\pi_1, \ldots, \pi_n),\ 0 < \pi_i < \infty$. It is assumed that the weights have been designed so that $\pi_i$ is inversely proportional to the likelihood that the survey design selects an observation with the same demographic characteristics as observation $x_i$. That is, observations with a lower probability of being selected than they would have under a simple random sampling approach are assigned greater weight than they would receive in a simple random sampling scenario. Conversely, observations with a higher probability of selection receive lower weights than they would in a simple random sampling setup. The $\pi_i$-s are scaled to ensure that $\sum_{i=1}^n \pi_i = n$.

\subsection{Related Works}
Pseudo maximum likelihood (PMLE) based approaches are very popular to conduct parametric inference with complex survey data, where we posit a parametric model $f_\theta$ to model $\mathcal{P}$ and $\theta$ encodes the features of interest of $\mathcal{P}$. The pseudo loglikelihood of $\theta$ takes the form $\mathcal{L}(\theta) = \sum_{i=1}^n \pi_i \log f_\theta(x_i)$ \citep{WOOLDRIDGE20071281, anzstat_cxsurvey}. The pseudo likelihood estimate of $\hat{\theta}_{\rm PMLE}$ satisfies the first order condition $\frac{\partial \mathcal{L}(\theta)}{\partial \theta} =\sum_{i=1}^n \pi_i \frac{\partial}{\partial \theta} \log f_\theta(x_i)$. Under certain regularity condition \citep{5805f73c-4dfa-385e-bd6d-68424fb9f5be},
\begin{align*}
    \sqrt{n}(\hat{\theta}_{\rm PMLE} - \theta_0) \overset{\text{d}}{\to} \mbox{N}(0, H^{-1}_\pi V_\pi H^{-1}_\pi),
\end{align*}
where $\theta_0$ is the true value of $\theta$, and $H_\pi$ and $V_\pi$ are estimated by 
\begin{align*}
   &\hat{H}_\pi = \frac{1}{n}\sum_{i=1}^n \pi_i \frac{\partial^2}{\partial \theta \partial \theta^{T}} \log f_\theta(x_i)\big|_{\theta = \hat{\theta}_{\rm PMLE} },\\
   & \hat{V}_\pi = \frac{1}{n}\sum_{i=1}^n \pi_i \frac{\partial \log f_\theta(x_i)}{\partial \theta}  \frac{\partial \log f_\theta(x_i)}{\partial \theta^{T}}\big|_{\theta = \hat{\theta}_{\rm PMLE} }
\end{align*}
respectively.

As an alternative, a semi-parametric inference framework can be developed where the feature of interest $\theta$ of the finite population $\mathcal{P}:=\{X_i,\ i\in[N]\}$, instead a of parametric family of distributions as earlier, is described by the set of estimating equations $\frac{1}{N}\sum_{i=1}^N g(X_i, \theta) = 0$ with a vector known functions $g$. This approach avoids complete parametric specification of the model, and widely utilized in Statistics and Econometrics \citep{Chib2018, chib2021bayesian}. Given a sample  $x = (x_1,\ldots, x_n)^\T$ and survey weights $x = (\pi_1,\ldots, \pi_n)^\T$, the exponentially tilted empirical likelihood \citep{10.1093/biomet/92.1.31} is given by 
\begin{align*}
    L_{\rm MCM}(\theta) =\big\{\prod_{i=1}^n w^{\star}_{i}:{w}^{\star} = \argmax_{w} \mbox{H}_n(\mathbf{w}),\ \mathbf{w}\in\mathcal{S}_{n-1}, \ \sum_{i=1}^n w_i [\pi_i g(x_i, \theta)] = 0  \big\}.
\end{align*}
Here and elsewhere, we use MCM as an acronym for {\em moment condition model}.  When the convex hull of $\cup_{i=1}^n g(x_i, \theta)$ contains the origin, leading to $L_{\rm MCM}(\theta) = \prod_{i=1}^n w_i^\star(\theta)$, with 
\begin{align*}
    w_{i}^{\star}(\theta) =  \frac{\exp[\pi_i\lambda(\theta)^{\T} g(x_i, \theta)]}{\sum_{j=1}^n \exp[\pi_j\lambda(\theta)^{\T} g(x_j, \theta)]}
\end{align*}
and $\lambda(\theta) = \argmin_{\eta} n^{-1} \sum_{i=1}^n \exp[\pi_i\eta^{\T} g(x_i, \theta)]$.
When the convex hull condition is not satisfied, $L_{\rm MCM}(\theta\mid x_1,\ldots,x_n)$ is set to zero. 

\subsection{Proposed Methodology}
Importantly, it is often unwieldy, if not impossible, to put more flexible constraints on the parameter of interest via moment conditions. In this article, we intend to provide the additional flexibility  to the ETEL framework via providing the scope for statistical distance based parametric  distribution guided constraints.  However, it is not straight forward to accomplish that in the context of complex survey data, due to the presence of the survey weights. To carefully circumnavigate this issue,  we first reconstruct $M$  pseudo true populations of size $N$ from the observed complex survey sample of size $n$ via Weighted Finite Population Bayesian bootstrap \citep{dong2014a, cohen1997bayesian,  lo1993bayesian} to adjust for the survey weights; next draw an i.i.d pseudo sample of size $n$ from each of the pseudo true populations, and finally construct an ETEL based on each of the  $M$ pseudo samples. Given the $m$-th i.i.d pseudo sample $(x^{\star}_{m,1},\ldots, x^{\star}_{m, n}), \ m \in[M]$, the exponentially tilted empirical likelihood with parametric distribution guided constraint takes the form 
\begin{align*}
L_{\rm BDCM}(\theta) =&\bigg\{\prod_{i=1}^n w^{\star}_{i}:{w}^{\star} = \argmax_{w} \mbox{H}_n(\mathbf{w}),\ \mathbf{w}\in\mathcal{S}_{n-1},\notag\\
&\sum_{i=1}^n w_i g(x^{\star}_{m, i}, \theta) = 0,\
W_{2}^2\bigg[\sum_{i=1}^n w_i \delta_{x^{\star}_i}(\cdot),\ f_{\theta}\bigg] \leq \varepsilon  \bigg\},
\end{align*} 
where $\delta$ is the indicator function, $f_\theta$ is a parametric distribution of choice, and $\varepsilon$ is an user defined parameter denoting the maximum extent of departure from the parametric distribution of choice. Here and elsewhere, we use BDCM as an acronym for {\em bootstrapped distributionally constrained models}. Importantly, the inference on the $M$ pseudo true samples can be carried out in parallel. The final estimates of $\theta$ is obtained via combining the estimated obtained from the $M$ i.i.d pseudo samples.

\begin{table}
\begin{center}
\begin{tabular}{c c c c c} 
 \hline
$n$ & $\rho$ & 0.1 & 0.5 & 0.8 \\ 
\hline
500&  MLE &0.19 (0.91)&0.68  (0.48)&1.11 (0.45)\\
&  BPPE & 0.67 (0.82) & 0.65 (0.72) & 0.71 (0.78) \\ 
&  PMLE & 0.16 (0.94)& 0.16 (0.91)& 0.16 (0.94) \\ 
 &BDCM  & 0.16 (0.92)&0.16 (0.93) &0.16 (0.95)
\\ 
 \hline 
1000&  MLE & 0.16(0.87)&0.69(0.48)& 1.11(0.42) \\
&  BPPE &0.15 (0.92)& 0.18 (0.90) & 0.18(0.92)\\ 
    & PMLE &  0.11 (0.94) &0.10 (0.94) &0.10 (0.92)\\ 
 &BDCM  &0.11 (0.93) &0.10 (0.94) &0.10 (0.96)\\ 
 \hline 
1500&  MLE & 0.15(0.84)& 0.68(0.47)& 1.11(0.42) \\
&  BPPE & 0.12 (0.94)& 0.10 (0.89)& 0.12 (0.90)
 \\ 
    & PMLE  & 0.09(0.94)& 0.08(0.94) &0.07 (0.93) \\ 
 &BDCM  &0.09(0.94) &0.08(0.93)& 0.08(0.94)\\ 
\hline
2000&  MLE & 0.15 (0.81)& 0.68 (0.48)& 1.10 (0.40)\\
&  BPPE &0.09 (0.92)& 0.08 (0.92)& 0.07(0.92)
\\ 
    & PMLE  &  0.07(0.95) &0.07(0.95) &0.06(0.92) \\ 
    & BDCM   &0.07 (0.95) &0.07 (0.97)& 0.07 (0.97)\\ 
\hline
2500&  MLE & 0.15 (0.75) &0.68 (0.47)& 1.10 (0.39)\\
&  BPPE &0.06(0.94)& 0.07(0.88) &0.06 (0.92) \\ 
    & PMLE  & 0.06(0.96) &0.07 (0.94)& 0.06 (0.92)\\ 
 &BDCM   &0.06(0.97) &0.06 (0.95) &0.06 (0.94)\\ 
\hline
\end{tabular}
\caption{\label{cx_sim1_bias} Average bias (=$||(\mu, \sigma^2) - (\hat{\mu}, \hat{\sigma^2})||$) and coverage (within braces) of the MLE, PMLE,  BPPE \citep{leon-novelo2019fully}, BDCM estimators for varying data generating mechanisms.}
\end{center}
\end{table}

\subsection{Experiments}
Based on the numerical experiments in \citep{anzstat_cxsurvey}, we design simulation studies to compare the proposed distribution guided guided entropy maximization approach with the popular pseudo likelihood approach.  Suppose the random variables $(X, Z)$ jointly follows a bivariate normal distribution  with mean $(\mu_x, \mu_z)^\prime = (0, 10)^\prime$, marginal variances $(\sigma^2_x, \sigma^2_z)^\prime = (4, 16)^\prime$ and correlation $\rho\in\{0.1, 0.5, 0.8\}$. The variable $X$ is the variable of interest; we aim to estimate its mean $\mu_x$ and variance $\sigma^2_x$. The variable $Z$ is a selection variable, i.e the
$Z$-value a population unit determines the probability of inclusion of the unit into the sample. Particularly, we posit that the inclusion probability of  $X_s$  into the sample is given by 
$\pi^{\star}_s =  \Phi(\beta_0 + \beta_1 Z_s),  
$, where $\Phi(\cdot)$ is the cumulative distribution function of a standard normal distribution. When a population unit is included in the sample, we observe $x_s$ and assign a  survey weight $\pi_s$ such that $\pi_s \propto 1/\pi^{\star}_s$. Importantly, we assume that we do not directly observe $Z_s$. The selected sample  of size $n$ is denoted as $(\mathbf{x}, \mathbf{\pi)^\prime}$. We scale the weights such that they sum up to $1$, and  we have 
$\pi_s = \frac{(1/\pi_s^{\star})}{\sum_{j=1}^n  (1/\pi_j^{\star})}, \quad s \in [n]$. The objective is to utilize $(\mathbf{x}, \mathbf{\pi)^\prime}$ to estimate the population parameters of interest $(\mu_x, \sigma^2_x)$. 

We generate $N = 100, 000$ values of $(X_s, Z_s)$ 
as a finite population. We set $\beta_0 = 0.1,\ \beta_1 = -1.8$, and draw samples of sizes $n\in\{500, 1000, 1500, 2000, 2500\}$ from the finite population. Under each data generating set up, we utilize $100$ Monte Carlo simulations. For the Pseudo maximum likelihood (PMLE) approach, we simply posit the model $f_\theta\equiv \mbox{Normal}(\mu_x, \sigma^2_x)$.
For the proposed BCDM approach, we assume the moment constraint based on the function $g(x, \mu_x) = x -\mu_x$, and the  Wasserstein distance constraint based on the parametric family of distributions $f_\theta\equiv \mbox{Normal}(\mu_x, \sigma^2_x)$.  For each of the replicates, we choose $M=500$; and to ensure comparability of PMLE and BDCM, we set $\varepsilon = W_{2}^2\big[\sum_{i=1}^n 1/n \delta_{x^{\star}_i}(\cdot),\ f_{\hat{\theta}}\big]$, where $\hat{\theta}$ is the estimate of $\theta$ obtained via PMLE. The bias and the coverage of the pseudo maximum likelihood and moment condition model based estimators for varying data generating mechanisms are presented in Table \ref{cx_sim1_bias}. A case study with complex survey data from National Health and Nutrition Examination Surveys (NHANES) is provided in the supplement.

\subsection{National Health and Nutrition Examination Surveys (NHANES) Data Analysis}
NHANES is a series of surveys designed to assess the health and nutritional status of individuals in the United States. The data extracted is from NHANES 2009-2010 \citep{nhanes} that contains information on binary indicator of high cholesterol, race, age etc., and survey weights for $8591$ individuals. For this exercise, we assume that these $8591$ individuals make up a finite population, and obtain samples of size $n\in\{250, 500, 1000, 2000\}$ according to the survey weights, and fit a logistic regression to model the binary indicator of high cholesterol as a function of race and age category.  For each $n\in\{250, 500, 1000, 2000\}$, we utilize $100$ Monte Carlo simulations. For the distribution guided entropy maximization approach, we assume  constraints on the score function  of the Logistic regression. The coverage of the  moment condition model based estimates of the regression coefficients is presented in Table \ref{cx_nhanes_bias}.

\begin{table}
\begin{center}
\begin{tabular}{ ccccc } 
 \hline
 $n$ & 250 & 500 & 1000 &  2000 \\ 
 \hline
 Coverage & 0.95 & 0.95 & 0.96 & 0.96 \\ 
 \hline
 Bias  & 0.42 & 0.27 & 0.18 & 0.13 \\ 
 \hline 
\end{tabular}
\caption{\label{cx_nhanes_bias} \textbf{NHANES Data.} Bias $ = ||\beta -\hat{\beta}||$ and coverage  of the moment condition model based estimates of the regression parameters for varying sample sizes.}
\end{center}
\end{table}

\section{Demographic Parity }\label{ssec:algorithmic_fairness}
Discrimination pertains to unfair treatment of individuals based on specific demographic characteristics known as protected attributes.  The goal of demographic parity or statistical parity \citep{90450a4b5b49471b8111fc88355f2e7f, gajane2018formalizing} in machine learning is to design algorithms that yield fair inferences devoid of discrimination due to membership to certain demographic groups determined by a protected attribute. First, we introduce the mathematical formalization of the notions of demographic parity. To that end,  we assume that $X$ denotes the feature vector used for predictions, $A$ is the protected attribute with two levels $\{S,T\}$, and  $Y$ is the response. Parity constraints are phrased in terms of the distribution over $(X, A, Y)$. Two definitions are in order.

\begin{definition}[Demographic parity, \citep{90450a4b5b49471b8111fc88355f2e7f}]\label{def:DP}
A predictor $h$  satisfies demographic parity under the distribution over  $(X, A, Y)$ if $h(X)$ is independent of the protected attribute $A$, i.e ,$\mathbf{P}[h(X) \geq z \mid A = S] = \mathbf{P}[h(X) \geq z \mid A = T] = \mathbf{P}[h(X) \geq z ]$,
for all $z$.
\end{definition}

\begin{definition}[Demographic parity in expectation, \citep{90450a4b5b49471b8111fc88355f2e7f}
]\label{def:DP_exp}
A predictor $h$  satisfies demographic parity under the distribution over  $(X, A, Y)$ if $h(X)$ is independent of the protected attribute $A$, i.e ,
$\mathbf{E}[h(X) \mid A = S] = \mathbf{E}[h(X)  \mid A = T] = \mathbf{E}[h(X) ].$
\end{definition}
\subsection{Proposed Methodology}
Although the notions of demographic parity in Definitions  \ref{def:DP} and \ref{def:DP_exp}  coincide when we work with binary responses, the latter may be amenable to simple computational algorithms \citep{e21080741} compared to the general definition. However, the notion of demographic parity in expectation is somewhat prohibitive since one cannot control the predictor $h$ over its entire domain. For example, depending on the application of interest, we may be solely interested in controlling the tails of the predictor \citep{yang2019fair}. Taking refuge to our semi-parametric inference framework, we offer a flexible as well as a computationally feasible compromise between the notions in Definitions \ref{def:DP} and \ref{def:DP_exp}. To that end, we introduce the notion of demographic parity in the Wasserstein metric next.
\begin{definition}[Demographic parity in Wasserstein metric]
A predictor $h$ achieves demographic parity in Wasserstein metric  with bias $\varepsilon$, under the distribution over  $(X, A, Y)$  if 
$W_2^{2}\big[F_{h_{S}}, F_{h_{T}}\big] \leq \varepsilon$,
where $F_{h_{k}}$ is the empirical distribution of $h$ under sub-population $k$ i.e $h(X)\mid A = k, k \in\{S,T\}$.
\end{definition}
Suppose we have data $(x_i, y_i, a_i)\in \mathbf{R}^d\times\mathbf{R}\times\{S,T\}$ for  $n$ individuals on $p$-dimensional covariate $x$, univariate continuous response $y$, and levels of the protected attribute $a\in\{S,T\}$.
For the sake of simplicity in exposition, we also assume that $a_i = S,\ i\in[n_S]$  and $a_i = T,\ i\in[n]\setminus[n_S]$ where $n = n_S + n_T$. Next, we posit a predictive model
    $y_i = h(x_i, \theta_{(a_i)}) + e_i, \ e_i \overset{i.i.d}\sim \mbox{N}(0, \sigma^2), \ i\in[n]$,
where $h$ is potentially non-linear, and $(\theta_{(S)}, \theta_{(T)}) $ is the model parameter of interest to be estimated under the demographic parity constraint  $W^{2}_{ 2}\big[F_{h_{S}}, F_{h_{T}}\big] \leq \varepsilon$. In particular, we consider the empirical cdf of $h$ under sub-population $S$,  $F_{h_{S}}= 1/n_S\sum_{i = 1}^{n_S} \delta_{h(x_{iS})}(\cdot)$; and a weighted empirical cdf of $h$ under sub-population $T$, $F_{h_{T}}= \sum_{i = n_S + 1}^{n} w_{i}\ \delta_{h(x_{iT})}(\cdot)$. Here $\delta$ is the Dirac delta measure. The goal is to infer about $(\theta_{(S)}, \theta_{(T)}, w) $ ensuring that demographic parity constraint i.e $F_{h_{S}},F_{h_{T}}$ are close  with respect $W^{2}_{2}$, at the same time the extent of re-weighting in $F_{h_{T}}$ is minimal i.e the entropy $-\sum_{i=n_S + 1}^n w_i\log w_i$ is close to the maximal entropy $\log n_T$.  A related idea in \citet{pmlr-v115-jiang20a} deals with $W_{1}$ constrained fair classification problems, but our approach of additionally re-weighting the observations offers more flexibility with possible ramifications in studying fairness in mis-specified models. 

We achieve this through an \emph{in-model} approach solving the optimization problem:
{\footnotesize
\begin{align}\label{wgf_simultenous}
   &\max_{w, \theta_{(S)}, \theta_{(T)}, \sigma^2}
   \bigg[-\frac{1}{n_S}\sum_{i=1}^{n_S} l_{i}(\theta_{(S)}\mid x_i) - \sum_{i=n_s + 1}^n w_i l_{i}(\theta_{(T)}\mid x_i)\notag \\
   &-(1-\lambda^{\star}) W_{2}^{2}\big[F_{h_{S}}, F_{h_{T}}\big] - \lambda^{\star} \sum_{i=n_s + 1}^{n} w_i\log w_i \bigg]
\end{align}
}
where $\sum_{i=n_s + 1}^{n} w_i = 1$ and $l_{i}(\theta_{(a_i)}\mid x_i) = (y_i - h(x_i, \theta_{(a_i)}))^2/2\sigma^2, \ i\in[n]$. For a resulting re-weighting vector $w^{\star} = (w^{\star}_{n_S + 1},\ldots,w^{\star}_{n})^{\prime}$, we can obtain fair prediction at a new $x\in T$ via a weighted kernel density estimate at $x$. As a competitor to the \emph{in-model} scheme, motivated by popular post-processing schemes to ensure fairness \citep{postprocess1, nandy2022achieving}, we utilize  \emph{two-step} procedure: \\
\textbf{Step 1:} We obtain model parameter estimates by $ (\hat{\theta}_{(S)},\ \hat{\theta}_{(T)} ,\ \hat{\sigma}^2)=$
{\footnotesize
\begin{equation}\label{wgf_twostep_1}
 \argmax_{\theta_{(S)},\ \theta_{(T)} ,\ \sigma^2} \bigg[ - \frac{1}{n_S}\sum_{i=1}^{n_S} l_{i}(\theta_{(S)}\mid x_i) - \frac{1}{n_T}\sum_{i=n_s + 1}^n l_{i}(\theta_{(T)}\mid x_i)   \bigg]
\end{equation}
}
followed by a post-processing step at $(\hat{\theta}_{(S)},\ \hat{\theta}_{(T)} ,\ \hat{\sigma}^2)$ to obtain $w^{\star}$  \\
\textbf{Step 2:} 
\begin{equation}\label{wgf_twostep_2}
  \argmax_{w} \bigg[ -(1-\lambda^{\star})  \  W_{2}^{2}\big[F_{h_{S}}, F_{h_{T}}\big] - \lambda^{\star}  \sum_{i=n_s +1}^n w_i\log w_i \bigg].  
\end{equation}

A case study on algorithmic mental health monitoring is provided next. An additional case study on algorithmic criminal risk assessment is also included.

\subsection{Distress Analysis Interview Corpus (DAIC)}\label{application:daic}
The Distress Analysis Interview Corpus (DAIC) \citep{gratch-etal-2014-distress} is a multi-modal clinical interview collection, accessible upon request via the \href{https://dcapswoz.ict.usc.edu/}{\textcolor{blue}{DAIC-WOZ}} website. Computer agents based on such clinical interviews  are deemed to be used for making mental health diagnosis in realtion to certain employment decisions, and concerns about the fairness of such tools with respect to the biological gender of the individuals are raised.
Specifically,  we focus on predicting the PHQ-8 score, that captures the individual's severity of depression, as a function of the individual's verbal signals during the clinical interviews, while biological gender serves as a protected attribute. 
In particular, the Fourier series analysis of the speech signal of the individuals yield verbal attributes of interest, that in turn could be potentially used in diagnosis of the individual's severity of depression.  Therefore, it is of interest to develop novel methods to produce predictions while avoiding disparate treatment on the basis of the biological genders. More precisely, we want to ensure that the demographic parity constraint is satisfied here, which in this context, simply dictates that the weighted empirical CDFs of biological gender-specific fitted PHQ-8 scores  are \emph{identical} or \emph{similar}.  

\begin{figure}[!htb]
    \centering
    {{\includegraphics[width=13cm, height = 4cm]{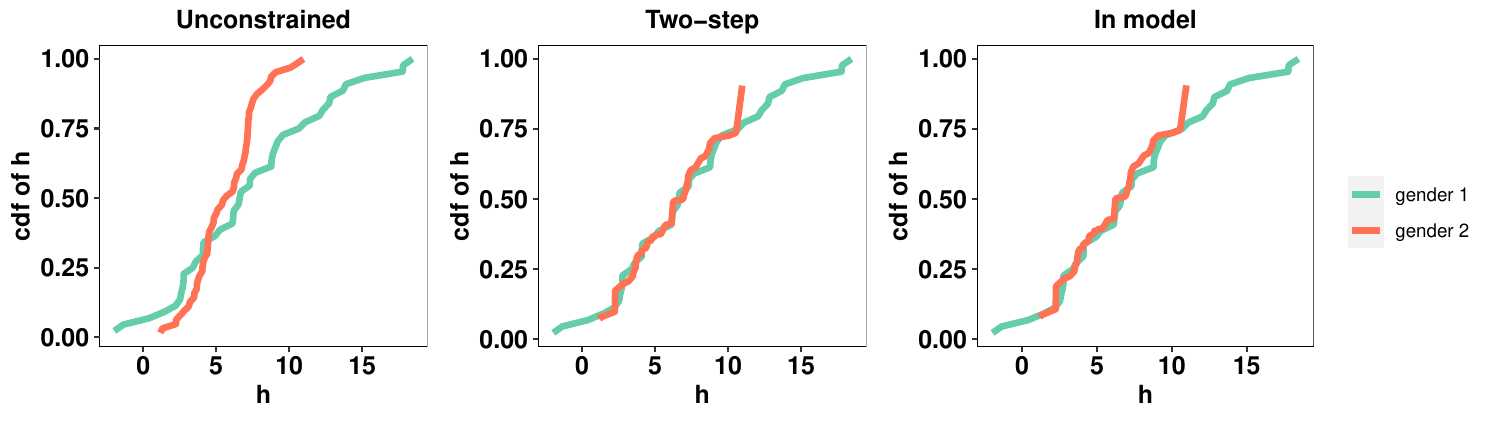} }}%
    
    \caption{\emph{\textbf{Distress Analysis Interview Corpus.} \emph{Empirical cdfs of  fitted $h$ for the two groups, with  no fairness constraint $( W_{2} = 19.32)$, fair post-processing $( W_{2} = 2.24)$, and fair model fitting with $( W_{2}= 0.79$) respectively at $\lambda^{\star} = 0$. }}\label{diag:daic_plots}}
\end{figure}

The PHQ-8 scores range from $0$ to $27$ with a score from $0-4$ considered none or minimal, $5-9$ mild, $10-14$ moderate, $15-19$ moderately severe, and $20-27$ severe. In this application, we work with this PHQ-8 (continuous response), biological gender (binary protected attribute), and $17$ derived audio/verbal features (continuous covariates) corresponding to the $n = 107$ subjects. The  PHQ-8 score for two biological genders show a clear discrepancy. Therefore, we shall assess the relative performance of the \emph{in-model} scheme in \eqref{wgf_simultenous}) and \emph{two-step} scheme \eqref{wgf_twostep_1}--\eqref{wgf_twostep_2} in ensuring demographic parity with respect to biological gender (refer to Figure \ref{diag:daic_plots}).  As earlier, for the sake of simplicity of exposition, we use linear regression (i.e $h$ is linear in the covariates) as our predictive model of choice.  When we fit the predictive model without any fairness constraint, the fitted empirical cumulative distribution functions corresponding to the two biological genders are widely different. Our \emph{in model} scheme, as well as \emph{two-step} scheme significantly reduce the discrepancy owing to their in-built fairness-based regularization. As noted earlier, the \emph{in model} scheme provides lower bias since it performs the two-step optimization simultaneously.

\begin{figure}[!htb]
    \centering
    {{\includegraphics[width=13cm, height = 4.5cm]{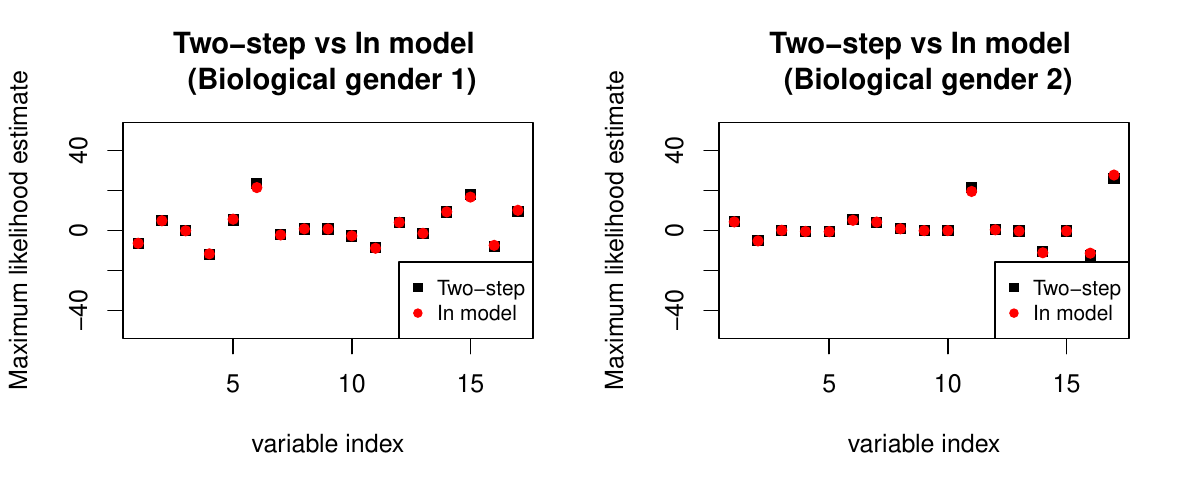} }}%
    
    \caption{\emph{\textbf{Distress Analysis Interview Corpus.} \emph{Maximum likelihood estimates of the regression coefficients under both two-step and in model schemes. In the in model scheme the estimates get slightly modified since the regression coefficients and the weights assigned to the data are learned simultaneously.  For details on the in model and two-step approaches, refer to equations \eqref{wgf_simultenous} and \eqref{wgf_twostep_1}-\eqref{wgf_twostep_2} respectively.}}\label{diag:daic_reg_plots}}
\end{figure}

\subsection{COMPAS Recidivism  Data Analysis}\label{sup:application:compas}
We consider a case study on algorithmic criminal risk assessment. We shall focus  on the popular COMPAS data set \citep{https://doi.org/10.1111/rssa.12613} that includes information on criminal history for the defendants in Broward County, Florida, available from the \href{https://www.propublica.org/datastore/dataset/compas-recidivism-risk-score-data-and-analysis}{\textcolor{blue}{propublica}} website.  For each individual, several features on criminal history are available, such as the number of past felonies, misdemeanors, and juvenile offenses; additional demographic information includes the sex, age, and ethnic group of each defendant.  We focus on predicting two-year recidivism score $y$ (continuous) as a function of the defendant’s demographic
information except for race and criminal history $x$, while race (categorical) serves as a protected attribute. Algorithms for making such
predictions are routinely used in courtrooms to advise judges, and concerns about the fairness of such
tools with respect to the race of the defendants are raised. Therefore, it is of interest to develop novel methods to produce predictions while avoiding disparate treatment on the basis of the protected attribute race. More precisely, we want to ensure that the demographic parity constraint is satisfied, which in this context, simply dictates that the weighted empirical CDFs of race-specific fitted recidivism scores  are \emph{identical} or \emph{similar}.  

\begin{figure}[!htb]
    \centering
    {{\includegraphics[width=13cm, height = 4cm]{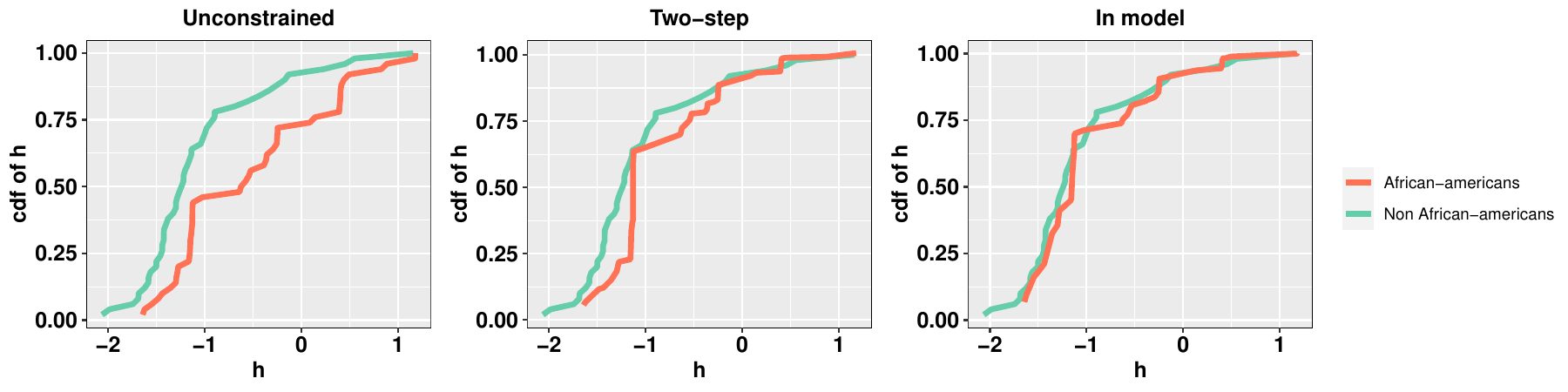} }}%
    
    \caption{\emph{\textbf{COMPAS dataset.} Empirical cdfs of  fitted $h$ for the two groups, with  no fairness constraint $( W_{2} = 0.72)$, fair post-processing $( W_{2} = 0.05)$, and fair model fitting with $( W_{2}= 0.02)$ respectively at $\lambda^{\star} = 0$.}\label{diag:compas_plots}}
\end{figure}
\begin{figure}[!htb]
    \centering
    {{\includegraphics[width=13cm, height = 4.5cm]{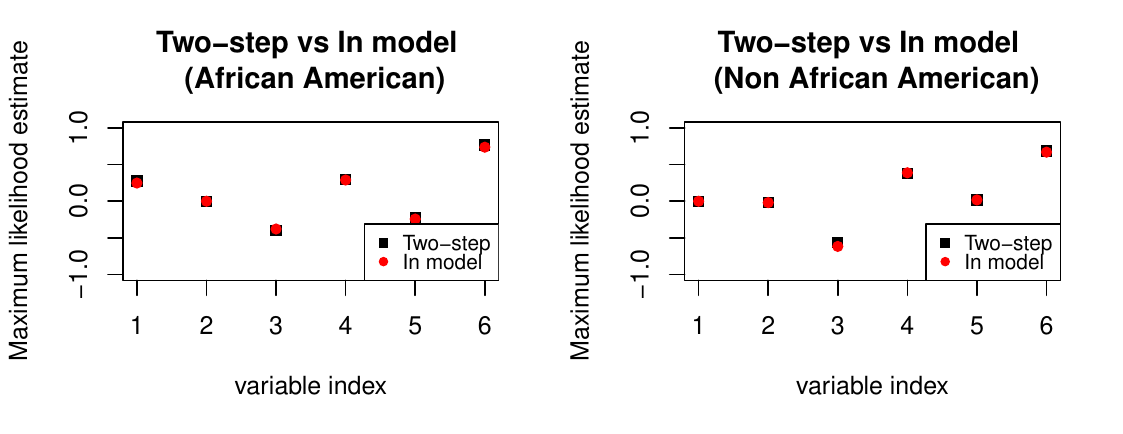} }}%
    
    \caption{\emph{\textbf{COMPAS dataset.} \emph{Maximum likelihood estimates of the regression coefficients under both two-step and in model schemes. In the in model scheme the estimates get slightly modified since the regression coefficients and the weights assigned to the data are learned simultaneously.  
    }}\label{diag:daic_reg_plots}}
\end{figure}   
For simplicity of exposition, we only consider two levels for the protected attribute race, namely, African-American or non-African-American, and consider a sub-sample of the entire data set with $100$ defendants corresponding to each level of the protected attribute. As covariate, for each defendant, we consider demographic information -- sex (binary), age (continuous), marital status (categorical); and criminal status -- legal status (categorical), supervision level (categorical), custody status (categorical). We use linear regression (i.e $h$ is linear in the covariates) as our predictive model of choice; the methodology readily extends to more complicated models. The histograms of raw recidivism score for African-Americans versus non-African-Americans show a clear discrepancy (refer to Figure \ref{diag:compas_data}). We shall assess the relative performance of the \emph{in-model} scheme in \eqref{wgf_simultenous} and \emph{two-step} scheme in \eqref{wgf_twostep_1}--\eqref{wgf_twostep_2} in ensuring demographic parity with respect to the protected attribute race (refer to Figure \ref{diag:compas_plots}). When we fit the predictive model without any fairness constraint, the fitted empirical cumulative distribution functions corresponding to the two sub-populations
are widely different. Our \emph{in-model} scheme, as well as \emph{two-step} scheme significantly reduce the discrepancy owing to their in-built fairness-based regularization. As expected, the \emph{in-model} scheme provides slightly lower bias since it performs the two-step optimization simultaneously.

\section{Entropy Based Portfolio Allocation}\label{ssec:po}
We present an application of the proposed parametric distribution guided entropy maximization framework to portfolio allocation problems \citep{10.2307/2975974, doi:10.1080/07474930801960394, elton2014modern}. Portfolio optimization is concerned with the allocation of an investor's wealth over several assets to optimize specific objective(s) based on historical data on asset returns. To elucidate the problem clearly, let $R_{(i)} = (R_{i,1} , R_{i,2},\ldots , R_{i, d} )^{\prime}$ be the excess returns on $d$ risky assets recorded over time  $i\in[n]$. The portfolio $(w_1, \ldots, w_d)$ is a vector of weights that represents the investor’s relative allocation of the wealth satisfying  $\sum_{i=1}^d w_i = 1$ and $w_i \geq 0, i\in[d]$. The goal is to learn the $(w_1, \ldots, w_d)$ subject to specific constraints based on historical data.

\subsection{Related Works}
Markowitz's mean-variance optimization \citep{10.2307/2975974} is widely recognized as one of the foundational formulations of the portfolio selection problem. The traditional mean variance (MV) optimal portfolio weights \citep{10.2307/2975974} are obtained via
\begin{align*}
  \mbox{argmax}_{w}\big[ w^\T\mu - \frac{\lambda}{2} w^\T\Sigma w \big],  
\end{align*}
such that $\sum_{i=1}^d w_i = 1$, where $\mu =(\mu_1, \ldots, \mu_d)^\T = (1/n) \sum_{i=1}^n R_{(i)}$ and $\Sigma = (1/n) \sum_{i=1}^n (R_{(i)} -\mu)(R_{(i)} -\mu)^{\T}$ are the mean and variance of the historical return, and  $\lambda>0$ is a risk aversion parameter. Given a specific mean and covariance matrix, the Markowitz paradigm offers an elegant approach to achieve an efficient allocation where the pursuit of higher expected returns inevitably entails assuming greater risk. However,  in this framework, it is essential either for the asset returns to follow a normal distribution or for the utility to solely depend on the first two moments. Real-world financial returns, as indicated by empirical evidence \citep{mills1995modelling, peiro1999skewness}, diverge from normal distribution assumptions and commonly exhibit heavier tails and  lack of  symmetry. To that end, \citep{MEHLAWAT2021348, doi:10.1080/14697681003756877} proposed to utilize higher order moments in the portfolio allocation problem. However, portfolios created using sample moments of stock returns tend to be excessively concentrated in a small number of assets, which contradicts the fundamental principle of diversification. To that end, several approaches are  proposed in the literature that ensures shrinkage of the portfolio weights towards maximum
diversification  \citep{doi:10.1080/07474930801960394, zhou2015portfolio, KANG2021126260}, i.e maximizes the entropy of the portfolio weights. In particular, \citep{doi:10.1080/07474930801960394} proposed to obtain the portfolio weights solving the optimization problem $\argmax_{w} \mbox{H}_d(\mathbf{w})$ subject to  $\sum_{i=1}^d w_i\mu_i \geq \mu_0, \  w^\T\Sigma w \leq \sigma^{2}_0$, such that $\sum_{i=1}^d w_i = 1$, and $(\mu_0, \sigma^{2}_0)$ are the target mean and variance of the portfolio return. Basically, this approach constitutes obtaining the portfolio weight via entropy maximization subject to moment based constraints.


\subsection{Proposed Methodology}
Importantly, empirical evidence suggests that, there is  merit in modeling the asset returns via non-normal distributions \citep{doi:10.1080/14697681003756877, park2021finding}, e.g skew-normal distribution \citep{SkewNormal}.
However, it is often unwieldy to put more flexible constraints on the portfolio weights in terms of moment conditions. In this section, we intend to provide the additional flexibility  to the entropy based portfolio optimization framework via providing the scope for statistical distance based parametric  distribution guided constraints.
Our semi-parametric framework provides an formidable alternative to the existing literature, since (a) we can  flexibly specify the distribution of the expected return, and (b) the entropy provides direct handle on portfolio diversity. We achieve this by obtaining portfolio weights via the optimization problem 
$\argmax_{w} \mbox{H}_d(\mathbf{w})$ subject to $W_2^{2}\big[\frac{1}{n}\sum_{i=1}^T\delta_{w^{T}R_{(i)}}(\cdot), f_{\theta_{0}}\big]\leq \varepsilon$, such that $\sum_{i=1}^d w_i = 1$. Here $\frac{1}{n}\sum_{i=1}^n\delta_{w^{T}R_{(i)}}(\cdot)$ is the empirical distribution of the portfolio return,  $f_{\theta}$ is the centering parametric family of distribution of choice, $\theta_0$ is the fixed target value of $\theta$, and $\varepsilon$ is user defined parameter. For practical purposes, it is useful to express the optimization problem above as the following
\begin{equation}\label{eqn:po_ours2}
\begin{split}
   \argmin_{w} &\bigg[ (1-\lambda^{\star})   W_{2}^{2}\bigg(\frac{1}{n}\sum_{i=1}^n\delta_{w^{T}R_{(i)}}(\cdot),\ f_{\theta_{0}}\bigg)
   -\lambda^{\star}\ b_d  \mbox{H}_d(\mathbf{w}) \bigg] \  
\end{split}  
\end{equation}
such that $\sum_{i=1}^d w_i = 1$ and $b_d = 1/\log d$. This choice of $b_d$ is convenient since it ensures that $b_d\mbox{H}_d(\mathbf{w})\in[0,1]$.  Further, the user defined parameter $\lambda^{\star}\in[0, 1]$ controls the balance between the portfolio diversity and extent of deviation from the target distribution $f_{\theta_{0}}$. 

For exposition in this article, we choose $f_{\theta_{0}}$ to be a Skew-normal distribution \citep{SkewNormal} with parameters $\theta = (\omega, \zeta, \alpha)^{\prime}$. 
For $\alpha=0$, we can recover the Normal distribution as absolute value of skewness increases and absolute value of $\alpha$ increases. For $\alpha>0$ the distribution left skewed and it is right skewed for $\alpha<0$. If $Z\sim\mbox{SN}(\zeta, \omega, \alpha)$, then we have $\mu_0 = \mathbf{E}(Z)
,
\sigma_{0}^2 = \mbox{Var}(Z) 
$, $\gamma_0 = \mbox{Skewness} 
$. This allows us to set $(\zeta, \omega,\alpha)$ to achieve  target $\theta_0 = (\mu_0,\sigma_{0}^2,\gamma_0)$ of the portfolio return distribution. This resulting skew normal density with fully specified parameters then serve as the target distribution to calculate  portfolio weights based on \eqref{eqn:po_ours2}.  The user can select any flexible probability distribution for modelling the portfolio return and follow the prescribed recipe to compute target parameter values.

\subsection{Historical Stock Returns Data Analysis}
\begin{figure}[!htb]
    \centering
    {\includegraphics[width=13cm, height = 4cm]{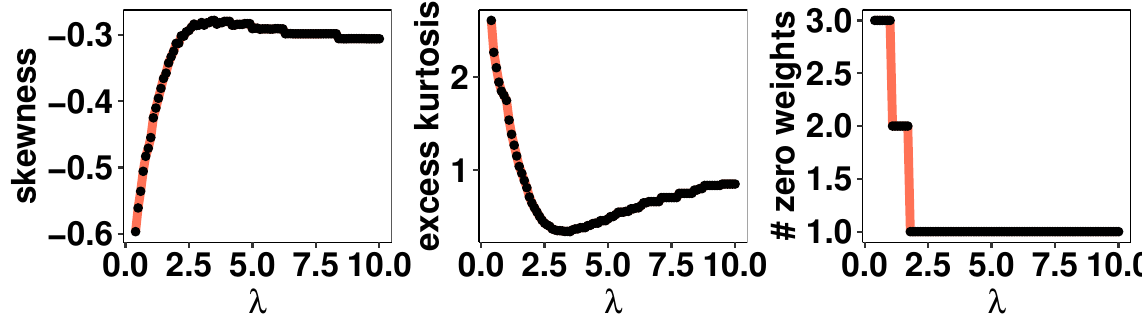} }%
    \caption{\emph{Limitations of mean-variance optimal portfolio: (i) The skewness and excess kurtosis plots provide evidence that the normality assumption for expected returns does not hold. (ii) Small value of $\lambda$ leads to zero weight to several assets}.\label{diag:MV}}
\end{figure}

\begin{figure}
\centering
    {\includegraphics[width=13cm, height = 4cm]{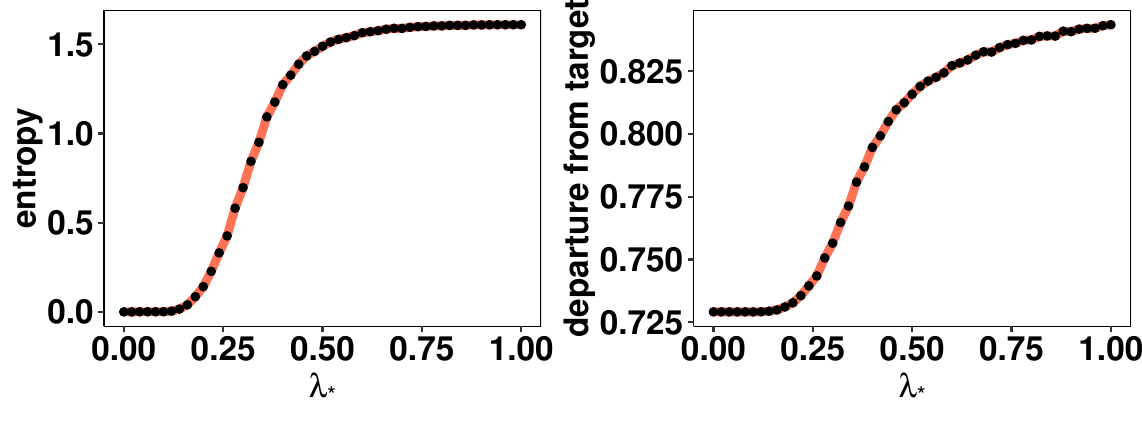} }
    \caption{\emph{With a fixed target skew normal return,  varying values of $\lambda^{\star}\in[0,1]$ provide different balances between diversity \& departure from target. Desired degree of diversification can be achieved $\lambda_{\star}$ via a simple grid search on $\lambda_{\star}\in[0, 1]$  .\label{diag:MEPO}}}%
\end{figure}

We consider stock returns data of $5$ companies (AMZN, AAPL, XOM, T,  MS) for the period January $2000$ to December $2020$, publicly available from Yahoo! Finance. The data is aggregated at monthly level. The goal is to compare mean-variance optimal portfolio and the proposed parametric distribution guided portfolio allocation frame work. First, we compute the mean-variance optimal portfolio   for varying value of the risk aversion parameter $\lambda\in[0, 10]$. Figure  \ref{diag:MV} records the skewness, excess kurtosis, and number of zero portfolio weights for the mean-variance optimal portfolio for varying $\lambda$. We focus on $\lambda$ set at $1$ -- a choice at which 3 out of 5 portfolio weights are 0, and the optimal portfolio return distribution is negatively skewed and leptokurtic. This exposes the fact that, once we have fixed the $\lambda$, mean-variance optimal portfolio optimization framework does not offer direct control over portfolio diversity, and we potentially obtain portfolio allocations concentrated on very few assets. Next, we fix the parameters of a skew-normal density $\theta = (\omega, \zeta, \alpha)^{\prime}$ such that it's mean, variance, and skewness match with the same quantities of the mean-variance optimal portfolio return at $\lambda = 1$. Finally, we compute the skew normal distribution guided maximum entropy portfolio, for varying value of the balance parameter $\lambda^{\star}\in[0,1]$ in \eqref{eqn:po_ours2}. Figure \ref{diag:MEPO} present the entropy of the portfolio weights and the departure of the portfolio return distribution from the guiding skew normal distribution as a function of  $\lambda^{\star}\in[0,1]$. This showcases that, contrary to the mean-variance optimal portfolio allocation, here the fund manager can choose a specific $\lambda^{\star}$ to ensure desired level of portfolio diversity, while maintaining fidelity towards a pre-specified  distribution of the portfolio return distribution.


\section{Concluding Remarks}\label{sec:conclusion}
We introduced a nonparametrically flavoured framework that aims to align the maximum entropy weight adjusted empirical distribution of observed data closely with a predefined and potentially continuous probability distribution, while permitting mild deviations. The framework's versatility is showcased in three distinct applications. We anticipate the proposed methodology's utility in numerous other statistical tasks requiring data re-weighting, e.g robustness \citep{wang2017robust},  covariate shifts \citep{wang2017robust}, ill-posed inverse problems \citep{gamboa1997bayesian}, etc.

\bibliography{paper-ref}

\begin{thebibliography}{53}
\providecommand{\natexlab}[1]{#1}
\providecommand{\url}[1]{\texttt{#1}}
\expandafter\ifx\csname urlstyle\endcsname\relax
  \providecommand{\doi}[1]{doi: #1}\else
  \providecommand{\doi}{doi: \begingroup \urlstyle{rm}\Url}\fi

\bibitem[Agarwal et~al.(2019)Agarwal, Dud{\'i}k, and Wu]{90450a4b5b49471b8111fc88355f2e7f}
Alekh Agarwal, Miroslav Dud{\'i}k, and {Zhiwei Steven} Wu.
\newblock Fair regression: Quantitative definitions and reduction-based algorithms.
\newblock In \emph{36th International Conference on Machine Learning, ICML 2019}, 36th International Conference on Machine Learning, ICML 2019, pages 166--183. International Machine Learning Society (IMLS), January 2019.
\newblock 36th International Conference on Machine Learning, ICML 2019 ; Conference date: 09-06-2019 Through 15-06-2019.

\bibitem[Aliverti et~al.(2021)Aliverti, Lum, Johndrow, and Dunson]{https://doi.org/10.1111/rssa.12613}
Emanuele Aliverti, Kristian Lum, James~E. Johndrow, and David~B. Dunson.
\newblock Removing the influence of group variables in high-dimensional predictive modelling.
\newblock \emph{Journal of the Royal Statistical Society: Series A (Statistics in Society)}, 184\penalty0 (3):\penalty0 791--811, 2021.
\newblock \doi{https://doi.org/10.1111/rssa.12613}.
\newblock URL \url{https://rss.onlinelibrary.wiley.com/doi/abs/10.1111/rssa.12613}.

\bibitem[Azzalini and Valle(1996)]{SkewNormal}
A.~Azzalini and A.~DALLA Valle.
\newblock {The multivariate skew-normal distribution}.
\newblock \emph{Biometrika}, 83\penalty0 (4):\penalty0 715--726, 12 1996.
\newblock ISSN 0006-3444.
\newblock \doi{10.1093/biomet/83.4.715}.
\newblock URL \url{https://doi.org/10.1093/biomet/83.4.715}.

\bibitem[Bera and Park(2008)]{doi:10.1080/07474930801960394}
Anil~K. Bera and Sung~Y. Park.
\newblock Optimal portfolio diversification using the maximum entropy principle.
\newblock \emph{Econometric Reviews}, 27\penalty0 (4-6):\penalty0 484--512, 2008.
\newblock \doi{10.1080/07474930801960394}.
\newblock URL \url{https://doi.org/10.1080/07474930801960394}.

\bibitem[Campbell R.~Harvey and Müller(2010)]{doi:10.1080/14697681003756877}
Merrill W.~Liechty Campbell R.~Harvey, John C.~Liechty and Peter Müller.
\newblock Portfolio selection with higher moments.
\newblock \emph{Quantitative Finance}, 10\penalty0 (5):\penalty0 469--485, 2010.
\newblock \doi{10.1080/14697681003756877}.
\newblock URL \url{https://doi.org/10.1080/14697681003756877}.

\bibitem[Chakraborty et~al.(2023)Chakraborty, Bhattacharya, and Pati]{chakraborty2023robust}
Abhisek Chakraborty, Anirban Bhattacharya, and Debdeep Pati.
\newblock Robust probabilistic inference via a constrained transport metric, 2023.

\bibitem[Chernozhukov et~al.(2023)Chernozhukov, Newey, and Santos]{Chernozhukov2023}
Victor Chernozhukov, Whitney~K. Newey, and Andres Santos.
\newblock Constrained conditional moment restriction models.
\newblock \emph{Econometrica}, --\penalty0 (--):\penalty0 --, 2023.
\newblock \doi{10.3982/ECTA13830}.
\newblock Published online: 17 March 2023.

\bibitem[Chib et~al.(2018)Chib, Shin, and Simoni]{Chib2018}
Siddhartha Chib, Minchul Shin, and Anna Simoni.
\newblock {B}ayesian estimation and comparison of moment condition models.
\newblock \emph{Journal of the American Statistical Association}, 113\penalty0 (524):\penalty0 1656--1668, 2018.
\newblock \doi{10.1080/01621459.2017.1358172}.
\newblock URL \url{https://doi.org/10.1080/01621459.2017.1358172}.

\bibitem[Chib et~al.(2021)Chib, Shin, and Simoni]{chib2021bayesian}
Siddhartha Chib, Minchul Shin, and Anna Simoni.
\newblock {B}ayesian estimation and comparison of conditional moment models, 2021.
\newblock URL \url{https://arxiv.org/abs/2110.13531}.

\bibitem[Cohen(1997)]{cohen1997bayesian}
Michael~P. Cohen.
\newblock The bayesian bootstrap and multiple imputation for unequal probability sample designs.
\newblock Technical report, National Center for Education Statistics, 555 New Jersey Avenue NW, Washington DC 20208-5654, 1997.

\bibitem[Cover and Thomas(2012)]{cover2012elements}
Thomas~M Cover and Joy~A Thomas.
\newblock \emph{Elements of Information Theory}.
\newblock Wiley, 2012.

\bibitem[Dong et~al.(2014)Dong, Elliott, and Raghunathan]{dong2014a}
Q.~Dong, M.R. Elliott, and T.E. Raghunathan.
\newblock A nonparametric method to generate synthetic populations to adjust for complex sampling design features.
\newblock \emph{Survey Methodology}, 40:\penalty0 29--46, 2014.

\bibitem[Einmahl et~al.(2008)Einmahl, Krajina, and Segers]{Bernoulli14-4-2008}
John H.~J. Einmahl, Andrea Krajina, and Johan Segers.
\newblock A method of moments estimator of tail dependence.
\newblock \emph{Bernoulli}, 14\penalty0 (4):\penalty0 1003--1026, 2008.
\newblock \doi{10.3150/08-BEJ130}.

\bibitem[Elton et~al.(2014)Elton, Gruber, Brown, and Goetzmann]{elton2014modern}
Edwin~J. Elton, Martin~J. Gruber, Stephen~J. Brown, and William~N. Goetzmann.
\newblock \emph{Modern Portfolio Theory and Investment Analysis}.
\newblock Wiley, 2014.
\newblock ISBN 978-1118469941.

\bibitem[Eysenbach and Levine(2021)]{DBLP:journals/corr/abs-2103-06257}
Benjamin Eysenbach and Sergey Levine.
\newblock Maximum entropy {RL} (provably) solves some robust {RL} problems.
\newblock \emph{CoRR}, abs/2103.06257, 2021.
\newblock URL \url{https://arxiv.org/abs/2103.06257}.

\bibitem[Fitzsimons et~al.(2019)Fitzsimons, Al~Ali, Osborne, and Roberts]{e21080741}
Jack Fitzsimons, AbdulRahman Al~Ali, Michael Osborne, and Stephen Roberts.
\newblock A general framework for fair regression.
\newblock \emph{Entropy}, 21\penalty0 (8), 2019.
\newblock ISSN 1099-4300.
\newblock \doi{10.3390/e21080741}.
\newblock URL \url{https://www.mdpi.com/1099-4300/21/8/741}.

\bibitem[Gajane and Pechenizkiy(2018)]{gajane2018formalizing}
Pratik Gajane and Mykola Pechenizkiy.
\newblock On formalizing fairness in prediction with machine learning, 2018.
\newblock URL \url{https://www.fatml.org/media/documents/formalizing_fairness_in_prediction_with_ml.pdf}.

\bibitem[Gamboa and Gassiat(1997)]{gamboa1997bayesian}
F~Gamboa and E~Gassiat.
\newblock Bayesian methods and maximum entropy for ill-posed inverse problems.
\newblock \emph{The Annals of Statistics}, 25\penalty0 (1):\penalty0 328--350, 1997.

\bibitem[Gratch et~al.(2014)Gratch, Artstein, Lucas, Stratou, Scherer, Nazarian, Wood, Boberg, DeVault, Marsella, Traum, Rizzo, and Morency]{gratch-etal-2014-distress}
Jonathan Gratch, Ron Artstein, Gale Lucas, Giota Stratou, Stefan Scherer, Angela Nazarian, Rachel Wood, Jill Boberg, David DeVault, Stacy Marsella, David Traum, Skip Rizzo, and Louis-Philippe Morency.
\newblock The distress analysis interview corpus of human and computer interviews.
\newblock In \emph{Proceedings of the Ninth International Conference on Language Resources and Evaluation ({LREC}'14)}, pages 3123--3128, Reykjavik, Iceland, May 2014. European Language Resources Association (ELRA).
\newblock URL \url{http://www.lrec-conf.org/proceedings/lrec2014/pdf/508_Paper.pdf}.

\bibitem[Gudivada(2018)]{gudivada2018computational}
Venkat~N. Gudivada.
\newblock Computational analysis and understanding of natural languages: Principles, methods and applications.
\newblock In \emph{Handbook of Statistics}. 2018.

\bibitem[Gunawan et~al.(2020)Gunawan, Panagiotelis, Griffiths, and Chotikapanich]{anzstat_cxsurvey}
David Gunawan, Anastasios Panagiotelis, William Griffiths, and Duangkamon Chotikapanich.
\newblock {Bayesian weighted inference from surveys}.
\newblock \emph{Australia and Newzealand Journal of Statistics}, 2020.
\newblock URL \url{https://doi.org/10.1111/anzs.12284}.

\bibitem[Hall(2005)]{hall2005generalized}
Alastair Hall.
\newblock \emph{Generalized Method of Moments}.
\newblock Oxford University Press, 2005.

\bibitem[Jaynes(1957)]{jaynes1957information}
E.~T. Jaynes.
\newblock Information theory and statistical mechanics.
\newblock \emph{Physical Review. Series II}, 106\penalty0 (4):\penalty0 620--630, 1957.
\newblock \doi{10.1103/PhysRev.106.620}.

\bibitem[Jiang et~al.(2020)Jiang, Pacchiano, Stepleton, Jiang, and Chiappa]{pmlr-v115-jiang20a}
Ray Jiang, Aldo Pacchiano, Tom Stepleton, Heinrich Jiang, and Silvia Chiappa.
\newblock {W}asserstein fair classification.
\newblock In Ryan~P. Adams and Vibhav Gogate, editors, \emph{Proceedings of The 35th Uncertainty in Artificial Intelligence Conference}, volume 115 of \emph{Proceedings of Machine Learning Research}, pages 862--872. PMLR, 22--25 Jul 2020.
\newblock URL \url{https://proceedings.mlr.press/v115/jiang20a.html}.

\bibitem[Kardar(2007)]{kardar2007statistical}
Mehran Kardar.
\newblock \emph{Statistical Physics of Particles}.
\newblock Cambridge University Press, 2007.

\bibitem[Le{\'o}n-Novelo and Savitsky(2019)]{leon-novelo2019fully}
Luis~G. Le{\'o}n-Novelo and Terrance~D. Savitsky.
\newblock Fully {Bayesian} estimation under informative sampling.
\newblock \emph{Electron. J. Statist.}, 13\penalty0 (1):\penalty0 1608--1645, 2019.
\newblock \doi{10.1214/19-EJS1538}.

\bibitem[li~Kang et~al.(2021)li~Kang, Tian, Chen, Zhao, fu~Li, and Wei]{KANG2021126260}
Yan li~Kang, Jing-Song Tian, Chen Chen, Gui-Yu Zhao, Yuan fu~Li, and Yu~Wei.
\newblock Entropy based robust portfolio.
\newblock \emph{Physica A: Statistical Mechanics and its Applications}, 583:\penalty0 126260, 2021.
\newblock ISSN 0378-4371.
\newblock \doi{https://doi.org/10.1016/j.physa.2021.126260}.
\newblock URL \url{https://www.sciencedirect.com/science/article/pii/S0378437121005331}.

\bibitem[Lo(1993)]{lo1993bayesian}
Andrew~Y Lo.
\newblock A bayesian method for weighted sampling.
\newblock \emph{The Annals of Statistics}, 21:\penalty0 2138--2148, 1993.

\bibitem[Lumley(2010)]{nhanes}
Thomas Lumley.
\newblock \emph{Complex Surveys: A Guide to Analysis Using R: A Guide to Analysis Using R}.
\newblock John Wiley and Sons, 2010.

\bibitem[Magrans~de Abril et~al.(2018)Magrans~de Abril, Doya, et~al.]{magrans2018connectivity}
Ildefons Magrans~de Abril, Kenji Doya, et~al.
\newblock Connectivity inference from neural recording data: Challenges, mathematical bases and research directions.
\newblock \emph{Neural Networks}, 2018.

\bibitem[Mandt et~al.(2016)Mandt, McInerney, Abrol, Ranganath, and Blei]{mandt2016variational}
Stephan Mandt, James McInerney, Fahim Abrol, Rajesh Ranganath, and David Blei.
\newblock Variational tempering.
\newblock In \emph{Artificial Intelligence and Statistics}, pages 704--712, 2016.

\bibitem[Markowitz(1952)]{10.2307/2975974}
Harry Markowitz.
\newblock Portfolio selection.
\newblock \emph{The Journal of Finance}, 7\penalty0 (1):\penalty0 77--91, 1952.
\newblock ISSN 00221082, 15406261.
\newblock URL \url{http://www.jstor.org/stable/2975974}.

\bibitem[Mehlawat et~al.(2021)Mehlawat, Gupta, and Khan]{MEHLAWAT2021348}
Mukesh~Kumar Mehlawat, Pankaj Gupta, and Ahmad~Zaman Khan.
\newblock Portfolio optimization using higher moments in an uncertain random environment.
\newblock \emph{Information Sciences}, 567:\penalty0 348--374, 2021.
\newblock ISSN 0020-0255.
\newblock \doi{https://doi.org/10.1016/j.ins.2021.03.019}.
\newblock URL \url{https://www.sciencedirect.com/science/article/pii/S0020025521002565}.

\bibitem[Mills(1995)]{mills1995modelling}
T.~C. Mills.
\newblock Modelling skewness and kurtosis in the london stock exchange ft-se index return distributions.
\newblock \emph{Statistician}, 44:\penalty0 323--332, 1995.

\bibitem[Nandy et~al.(2022)Nandy, DiCiccio, Venugopalan, Logan, Basu, and El~Karoui]{nandy2022achieving}
Preetam Nandy, Cyrus DiCiccio, Divya Venugopalan, Heloise Logan, Kinjal Basu, and Noureddine El~Karoui.
\newblock Achieving fairness via post-processing in web-scale recommender systems.
\newblock In \emph{Proceedings of the 2022 ACM Conference on Fairness, Accountability, and Transparency (FAccT '22)}, pages 715--725. ACM, June 2022.
\newblock \doi{10.1145/3531146.3533136}.

\bibitem[Panaretos and Zemel(2019)]{2019}
Victor~M. Panaretos and Yoav Zemel.
\newblock Statistical aspects of {W}asserstein distances.
\newblock \emph{Annual Review of Statistics and Its Application}, 6\penalty0 (1):\penalty0 405–431, Mar 2019.
\newblock ISSN 2326-831X.
\newblock \doi{10.1146/annurev-statistics-030718-104938}.
\newblock URL \url{http://dx.doi.org/10.1146/annurev-statistics-030718-104938}.

\bibitem[Park(2021)]{park2021finding}
Jungjun Park.
\newblock Finding bayesian optimal portfolios with skew-normal returns.
\newblock August 2021.
\newblock 48 Pages Posted: 24 Jul 2020, Last revised: 16 Aug 2021.

\bibitem[Peiro(1999)]{peiro1999skewness}
A.~Peiro.
\newblock Skewness in financial returns.
\newblock \emph{Journal of Banking \& Finance}, 23:\penalty0 847--862, 1999.

\bibitem[Rachev et~al.(2007)Rachev, Stoyanov, and Fabozzi]{rachev2007advanced}
Svetlozar~T. Rachev, Stoyan Stoyanov, and Frank~J. Fabozzi.
\newblock \emph{Advanced Stochastic Models, Risk Assessment, and Portfolio Optimization: The Ideal Risk, Uncertainty, and Performance Measures}.
\newblock John Wiley \& Sons, 2007.

\bibitem[Ramas et~al.(2022)Ramas, Le, Chen, Kumar, and Rottmann]{GarridoRamas2022}
Jose~Garrido Ramas, Thu Le, Bei Chen, Manoj Kumar, and Kay Rottmann.
\newblock Unsupervised training data reweighting for natural language understanding with local distribution approximation.
\newblock In \emph{EMNLP 2022}, 2022.
\newblock URL \url{https://www.amazon.science/publications/unsupervised-training-data-reweighting-for-natural-language-understanding-with-local-distribution-approximation}.

\bibitem[Santambrogio(2015)]{noauthororeditor}
Filippo Santambrogio.
\newblock Optimal transport for applied mathematicians. calculus of variations, pdes and modeling, 2015.
\newblock URL \url{https://www.math.u-psud.fr/~filippo/OTAM-cvgmt.pdf}.

\bibitem[Schennach(2005)]{10.1093/biomet/92.1.31}
Susanne~M. Schennach.
\newblock {Bayesian exponentially tilted empirical likelihood}.
\newblock \emph{Biometrika}, 92\penalty0 (1):\penalty0 31--46, 03 2005.
\newblock ISSN 0006-3444.
\newblock \doi{10.1093/biomet/92.1.31}.
\newblock URL \url{https://doi.org/10.1093/biomet/92.1.31}.

\bibitem[Shannon(1948)]{shannon1948mathematical}
Claude~E Shannon.
\newblock A mathematical theory of communication.
\newblock \emph{Bell System Technical Journal}, 27\penalty0 (3):\penalty0 379--423, 1948.

\bibitem[Skilling and Bryan(1984)]{skilling1984maximum}
J.~Skilling and R.~K. Bryan.
\newblock Maximum entropy image reconstruction: general algorithm.
\newblock \emph{Monthly Notices of the Royal Astronomical Society}, 211:\penalty0 111--124, 1984.

\bibitem[Villani(2003)]{Villani2003TopicsIO}
C{\'e}dric Villani.
\newblock Topics in optimal transportation.
\newblock \emph{American Mathematical Society}, 2003.
\newblock URL \url{https://www.math.ucla.edu/~wgangbo/Cedric-Villani.pdf}.

\bibitem[Wang et~al.(2017)Wang, Kucukelbir, and Blei]{wang2017robust}
Yixin Wang, Alp Kucukelbir, and David~M. Blei.
\newblock Robust probabilistic modeling with bayesian data reweighting.
\newblock In \emph{Proceedings of the 34th International Conference on Machine Learning - Volume 70}, pages 3646--3655. JMLR. org, August 2017.

\bibitem[Wen et~al.(2014)Wen, Yu, and Greiner]{wen2014robust}
Jie Wen, Chun-Nam~John Yu, and Russell Greiner.
\newblock Robust learning under uncertain test distributions: Relating covariate shift to model misspecification.
\newblock In \emph{Proceedings of the International Conference on Machine Learning (ICML)}, 2014.

\bibitem[White(1982)]{5805f73c-4dfa-385e-bd6d-68424fb9f5be}
Halbert White.
\newblock Maximum likelihood estimation of misspecified models.
\newblock \emph{Econometrica}, 50\penalty0 (1):\penalty0 1--25, 1982.
\newblock ISSN 00129682, 14680262.
\newblock URL \url{http://www.jstor.org/stable/1912526}.

\bibitem[Wooldridge(2007)]{WOOLDRIDGE20071281}
Jeffrey~M. Wooldridge.
\newblock Inverse probability weighted estimation for general missing data problems.
\newblock \emph{Journal of Econometrics}, 141\penalty0 (2):\penalty0 1281--1301, 2007.
\newblock ISSN 0304-4076.
\newblock \doi{https://doi.org/10.1016/j.jeconom.2007.02.002}.
\newblock URL \url{https://www.sciencedirect.com/science/article/pii/S0304407607000437}.

\bibitem[Xian et~al.(2023)Xian, Yin, and Zhao]{postprocess1}
Ruicheng Xian, Lang Yin, and Han Zhao.
\newblock Fair and optimal classification via post-processing.
\newblock In Andreas Krause, Emma Brunskill, Kyunghyun Cho, Barbara Engelhardt, Sivan Sabato, and Jonathan Scarlett, editors, \emph{Proceedings of the 40th International Conference on Machine Learning}, volume 202 of \emph{Proceedings of Machine Learning Research}, pages 37977--38012. PMLR, 23--29 Jul 2023.
\newblock URL \url{https://proceedings.mlr.press/v202/xian23b.html}.

\bibitem[Yan et~al.(2022)Yan, Seto, and Apostoloff]{yan2022forml}
Bobby Yan, Skyler Seto, and Nicholas Apostoloff.
\newblock Forml: Learning to reweight data for fairness, 2022.

\bibitem[Yang et~al.(2019)Yang, Lafferty, and Pollard]{yang2019fair}
Dana Yang, John Lafferty, and David Pollard.
\newblock Fair quantile regression, 2019.
\newblock URL \url{https://arxiv.org/abs/1907.08646}.

\bibitem[Zhou et~al.(2015)Zhou, Yang, Yu, et~al.]{zhou2015portfolio}
R.~Zhou, Z.~Yang, M.~Yu, et~al.
\newblock A portfolio optimization model based on information entropy and fuzzy time series.
\newblock \emph{Fuzzy Optimization and Decision Making}, 14:\penalty0 381--397, 2015.
\newblock \doi{10.1007/s10700-015-9206-8}.

\end{thebibliography}
\bibliographystyle{plainnat}

\end{document}